\documentclass[10pt,twocolumn,letterpaper]{article}
\pdfoutput=1

\usepackage{iccv}
\usepackage{times}
\usepackage{epsfig}
\usepackage{graphicx}
\usepackage{amsmath}
\usepackage{amssymb}

\usepackage[utf8]{inputenc}	

\usepackage{anyfontsize}

\usepackage[english]{babel}	

\usepackage{microtype}		

\usepackage{enumerate}		

\usepackage{times}			

\usepackage{amsmath,amssymb,amsthm}

\usepackage{graphicx} 

\usepackage{overpic}

\usepackage{pgf}

\usepackage{pgfplots}
\newlength\figureheight 
\newlength\figurewidth 

\usepackage{tikz,tkz-euclide,tkz-graph}
\usetikzlibrary{calc}
\usetkzobj{all}

\usepackage{xypic}

\usepackage{tabularx}

\usepackage{caption}
\usepackage{booktabs}		

\usepackage{multirow}

\usepackage[english]{varioref}

\usepackage{algorithmicx,algorithm,algpseudocode}

\newcommand{\bb}[1]{\boldsymbol{\mathrm{#1}}}

\newcommand*\samethanks[1][\value{footnote}]{\footnotemark[#1]}

\usepackage{xargs} 
\usepackage[colorinlistoftodos,prependcaption,textsize=tiny]{todonotes}
\newcommandx{\unsure}[2][1=]{\todo[linecolor=red,backgroundcolor=red!25,bordercolor=red,#1]{#2}}
\newcommandx{\change}[2][1=]{\todo[linecolor=blue,backgroundcolor=blue!25,bordercolor=blue,#1{#2}}
\newcommandx{\info}[2][1=]{\todo[linecolor=OliveGreen,backgroundcolor=OliveGreen!
25,bordercolor=OliveGreen,#1]{#2}}
\newcommandx{\improvement}[2][1=]{\todo[linecolor=Plum,backgroundcolor=Plum!
25,bordercolor=Plum,#1]{#2}}
\newcommandx{\thiswillnotshow}[2][1=]{\todo[disable,#1]{#2}}
\usepackage[colorlinks,bookmarks=false]{hyperref}

\iccvfinalcopy 

\ificcvfinal\fi

\title{Geodesic convolutional neural networks on Riemannian manifolds}

\author{
Jonathan Masci$^\dagger$\thanks{equal contribution} \quad Davide Boscaini$^\dagger$\samethanks[1] \quad Michael M. Bronstein$^\dagger$ \quad Pierre Vandergheynst$^\ddagger$\\
$^\dagger$USI, Lugano, Switzerland \quad $^\ddagger$EPFL, Lausanne, Switzerland
}

\begin{document}

\maketitle

\begin{abstract}
Feature descriptors play a crucial role in a wide range of geometry analysis and processing applications, including shape correspondence, retrieval, and segmentation. 
In this paper, we introduce Geodesic Convolutional Neural Networks (GCNN), a generalization of the convolutional networks (CNN) paradigm to non-Euclidean manifolds. 
Our construction is based on a local geodesic system of polar coordinates to extract ``patches", which are then passed through a cascade of filters and linear and non-linear operators. 
The coefficients of the filters and linear combination weights are optimization variables that are learned to minimize a task-specific cost function. 
We use GCNN to learn invariant shape features, allowing to achieve state-of-the-art performance in problems such as shape description, retrieval, and correspondence. 
\end{abstract}

\vspace{-0.3cm}
\section{Introduction}

Feature descriptors are ubiquitous tools in shape analysis. 
Broadly speaking, a \emph{local} feature descriptor assigns to each point on the shape a vector in some multi-dimensional descriptor space representing the local structure of the shape around that point.
A \emph{global} descriptor describes the whole shape. 
Local feature descriptors are used in higher-level tasks such as establishing correspondence between shapes \cite{FuncMaps}, shape retrieval \cite{bronstein2011shape}, or segmentation \cite{skraba2010persistence}. 
Global descriptors are often produced by aggregating local descriptors e.g. using the bag-of-features paradigm. 
Descriptor construction is largely application dependent, and one typically tries to make the descriptor discriminative (capture the structures that are important for a particular application, e.g. telling apart two classes of shapes), robust (invariant to some class of transformations or noise), compact (low dimensional), and computationally-efficient. 

\paragraph*{Previous work}
Early works on shape descriptors such as spin images \cite{johnson1999using}, shape distributions \cite{osada2002shape}, and integral volume descriptors \cite{manay2006integral} were based on \emph{extrinsic} structures that are invariant under Euclidean transformations.  
The following generation of shape descriptors used \emph{intrinsic} structures such as geodesic distances \cite{elad2003bending} that are preserved by isometric deformations. 
The success of image descriptors such as SIFT \cite{lowe2004distinctive}, HOG \cite{dalal2005histograms}, MSER \cite{matas2004robust}, and shape context \cite{belongie2000shape} has led to several generalizations thereof to non-Euclidean domains (see e.g. \cite{zaharescu2009surface,digne2010level,ISC}, respectively).
The works~\cite{coifman2006diffusion,levy2006laplace} on diffusion and spectral geometry have led to the emergence of intrinsic spectral shape descriptors 
that are \emph{dense} and isometry-invariant by construction.  
Notable examples in this family include heat kernel signatures (HKS) 
\cite{HKS1} and wave kernel signatures (WKS) \cite{WKS}. 

Arguing that in many cases it is hard to model invariance but rather easy to create examples of similar and dissimilar shapes, Litman and Bronstein \cite{LearnDesc} showed that HKS and WKS can be considered as particular parametric families of transfer functions applied to the Laplace-Beltrami operator eigenvalues and proposed to learn an optimal transfer function. 
Their work follows the recent trends in the image analysis domain, where hand-crafted descriptors are abandoned in favor of learning approaches. 
The past decade in computer vision research has witnessed the re-emergence of ``deep learning'' and in particular, convolutional neural network (CNN) techniques \cite{fukushima1980neocognitron,lecun1989backpropagation}, allowing to learn task-specific features from examples. 
CNNs achieve a breakthrough in performance in a wide range of  applications such as image classification \cite{Krizhevsky:2012}, segmentation \cite{Ciresan:2012f}, detection and localization \cite{Sermanet14,Simonyan14c} and annotation 
\cite{FangGISDDGHMPZZ14,KarpathyF14}.

Learning methods have only recently started penetrating into the 3D shape analysis community in problems such as shape correspondence  \cite{shotton2013real,rodoladense}, similarity \cite{kanezakilearning}, description \cite{LearnDesc,windheuseroptimal,cormansupervised}, and retrieval \cite{litman2014supervised}.
CNNs have been applied to 3D data in the very recent works \cite{wu2015,kalogerakis2015} using standard (Euclidean) CNN architectures applied to volumetric 2D views shape representations, making them unsuitable for deformable shapes.  
Intrinsic versions of CNNs that would allows dealing with shape deformations are difficult to formulate due to the lack of shift invariance on Riemannian manifolds; we are aware of two recent works in that direction \cite{Bruna,WFT2015}. 

\vspace{-0.3cm}
\paragraph*{Contribution}
In this paper, we propose Geodesic CNN (GCNN), an extension of the CNN paradigm to non-Euclidean manifolds based on local geodesic system of coordinates that are analogous to `patches' in images. 
Compared to previous works on non-Euclidean CNNs \cite{Bruna,WFT2015}, our model is generalizable (i.e., it can be trained on one set of shapes and then applied to another one), local, and allows to capture anisotropic structures. 
We show that HKS \cite{HKS1}, WKS \cite{WKS}, optimal spectral descriptors \cite{LearnDesc}, and intrinsic shape context \cite{ISC} can be obtained as particular configurations of GCNN; therefore, our approach is a generalization of previous popular descriptors.
Our experimental results show that our model can be applied to achieve state-of-the-art performance in a wide range of problems, including the construction of shape descriptors, retrieval, and correspondence. 

\section{Background}

We model a 3D shape as a connected smooth compact two-dimensional manifold (surface) $X$, possibly with a boundary $\partial X$. 
Locally around each point $x$ the manifold is homeomorphic to a two-dimensional Euclidean space referred to as the \emph{tangent plane} and denoted by $T_x X$. 
A \emph{Riemannian metric} is an inner product $\langle \cdot, \cdot \rangle_{T_x X} \colon T_x X \times T_x X \to \mathbb{R}$ on the tangent space depending smoothly on $x$.  

\paragraph*{Laplace-Beltrami operator (LBO)} 
is a positive semidefinite operator $\Delta_X f = - \text{div}(\nabla f)$, generalizing the classical Laplacian to non-Euclidean spaces. 
The LBO is \emph{intrinsic}, i.e., expressible entirely in terms of the Riemannian metric. 
As a result, it is invariant to isometric (metric-preserving) deformations of the manifold. 
On a compact manifold, the LBO admits an eigendecomposition $\Delta_X \phi_k = \lambda_k \phi_k$ with real eigenvalues $0 = \lambda_1 \leq \lambda_2 \leq \dots$. 
The corresponding eigenfunctions $\phi_1, \phi_2, \hdots$ form an orthonormal basis on $L^2(X)$, which is a generalization of the Fourier basis to non-Euclidean domains. 

\paragraph*{Heat diffusion on manifolds} 
is governed by the \emph{diffusion equation}, 
\begin{equation}
\label{eq:diffeq}
\left(\Delta_X + \tfrac{\partial}{\partial t}\right)u(x,t) = 0; \quad u(x,0) = u_0(x),
\end{equation}
where $u(x,t)$ denotes the amount of heat at point $x$ at time $t$, $u_0(x)$ is the initial heat distribution; if the manifold has a boundary, appropriate boundary conditions must be added. 
The solution of \eqref{eq:diffeq} is expressed in the spectral domain as 
\begin{equation}
\label{eq:heatop_}
u(x,t) = \int_X u_0(x') \underbrace{\sum_{k\geq 1}  e^{-t \lambda_k} \phi_k(x)\phi_k(x')}_{h_t(x,x')} dx',  
\end{equation}
where $h_t(x,x')$ is the \emph{heat kernel}. 
Interpreting the LBO eigenvalues as `frequencies', the coefficients $e^{-t\lambda}$ play the role of a transfer function corresponding to a low-pass filter sampled at $\{ \lambda_k\}_{k\geq 1}$. 

\paragraph*{Discretization}  
In the discrete setting, the surface $X$ is sampled at $N$ points $x_1, \hdots, x_N$. 
On these points, we construct a triangular mesh $(V,E,F)$ with vertices $V = \{1, \hdots, N\}$, in which each interior edge $ij \in E$ is shared by exactly two triangular faces $ikj$ and $jhi \in F$, and boundary edges belong to exactly one triangular face. 
The set of vertices $\{ j \in V : ij\in E \}$ directly connected to $i$ is called the \emph{1-ring} of $i$. 
A real-valued function $f \colon X \to \mathbb{R}$ on the surface is sampled on the vertices of the mesh and can be identified with an $N$-dimensional vector $\bb{f} = (f(x_1), \hdots, f(x_N))^\top$. 
The discrete version of the LBO is given as an $N\times N$ matrix $\bb{L} = \bb{A}^{-1}\bb{W}$, where 
\begin{equation}
w_{ij} = 
\begin{cases}
(\cot \alpha _{ij} + \cot \beta _{ij})/2 & ij \in E; \\
-\sum_{k\neq i} w_{ik} & i = j; \\
0 & \text{else}; 
\end{cases}
\label{eq:cotan}
\end{equation}
$\alpha_{ij}, \beta_{ij}$ denote the angles $\angle ikj, \angle jhi$ of the triangles sharing the edge $ij$, and $\bb{A} = \text{diag}(a_1, \hdots, a_N)$ with $a_i = \frac{1}{3} \sum_{jk: ijk \in F} A_{ijk}$ being the local area element at vertex $i$ and $A_{ijk}$ denoting the area of triangle $ijk$ \cite{Pinkall1993}. 

The first $K \leq N$ eigenfunctions and eigenvalues of the LBO are computed by performing the generalized eigendecomposition  $\bb{W} \boldsymbol{\Phi} = \bb{A}\boldsymbol{\Phi}\boldsymbol{\Lambda}$, where $\boldsymbol{\Phi} = (\boldsymbol{\phi}_1, \hdots, \boldsymbol{\phi}_K)$ is an $N\times K$ matrix containing as columns the discretized eigenfunctions and $\boldsymbol{\Lambda} = \text{diag}(\lambda_1, \hdots, \lambda_K)$ is the diagonal matrix of the corresponding eigenvalues. 

\section{Spectral descriptors}
Many popular spectral shape descriptors are constructed taking the diagonal values of heat-like operators. 
A generic descriptor of this kind has the form 
\begin{equation}
\mathbf{f}(x) = \sum_{k \geq 1} \boldsymbol{\tau}(\lambda_k) \phi^2_k(x) \approx \sum_{k=1}^K \boldsymbol{\tau}(\lambda_k) \phi^2_k(x)
\label{eq:gendesc}
\end{equation}
where $\boldsymbol{\tau}(\lambda) = (\tau_1(\lambda), \hdots, \tau_Q(\lambda))^\top$ is a bank of transfer functions acting on LBO eigenvalues, and $Q$ is the descriptor dimensionality. 
Such descriptors are dense (computed at every point $x$), intrinsic by construction, and typically can be efficiently computed using a small number $K$ of LBO eigenfunctions and eigenvalues. 

\paragraph*{Heat kernel signature (HKS)} 
\cite{HKS1} is a particular setting of~\eqref{eq:gendesc} using parametric low-pass filters of the form $\tau_t(\lambda) = e^{-t \lambda}$, which allows to interpret them as diagonal values of the heat kernel taken at some times $t_1, \hdots, t_Q$. 
The physical interpretation of the HKS is \emph{autodiffusivity}, i.e., the amount of heat remaining at point $x$ after time $t$, which is equal (up to constant) to the Gaussian curvature for small $t$. 
A notable drawback of HKS stemming from the use of low-pass filters is its poor spatial localization. 

\paragraph*{Wave kernel signature (HKS)} 
\cite{WKS} arises from the model of a quantum particle on the manifold possessing some initial  energy distribution, and boils down to a particular setting of~(\ref{eq:gendesc}) with band-pass filters of the form $\tau_\nu(\lambda) = \exp\left( \frac{\log \nu - \log \lambda}{2\sigma^2} \right)$, where $\nu$ is the initial mean energy of the particle. 
WKS have better localization, but at the same time tend to produce noisier matches. 

\paragraph*{Optimal spectral descriptors (OSD)} 
\cite{LearnDesc} use parametric transfer functions expressed as 
\begin{equation}
\label{eq:spline}
\tau_q(\lambda) = \sum_{m = 1}^M a_{qm} \beta_{m}(\lambda) 
\end{equation}
in the B-spline basis $\beta_1(\lambda), \hdots, \beta_M(\lambda)$, where $a_{qm}$ ($q=1,\hdots,Q, m=1,\hdots, M$) are the parametrization coefficients. 
Plugging~\eqref{eq:spline} into~\eqref{eq:gendesc} one can express the $q$th component of the spectral descriptor as 
\begin{eqnarray}
\label{eq:gendesc_}
\hspace{-2mm}
f_q(x) =\sum_{k \geq 1} \tau_q(\lambda_k) \phi^2_k(x) 
=  \sum_{m = 1}^M a_{qm}  \underbrace{ \sum_{k \geq 1}\beta_{m}(\lambda_k)  \phi^2_k(x) }_{g_m(x) },
\end{eqnarray}
where $\mathbf{g}(x) = (g_1(x), \hdots, g_M(x))^\top$ is a vector-valued function referred to as {\em geometry vector}, dependent only on the intrinsic geometry of the shape. 
Thus, \eqref{eq:gendesc} is parametrized by the $Q\times M$ matrix $\mathbf{A} = (a_{lm})$ and can be written in matrix form as $\mathbf{f}(x) = \mathbf{A}\mathbf{g}(x)$. 
The main idea of \cite{LearnDesc} is to \emph{learn} the optimal parameters $\mathbf{A}$ by minimizing a task-specific loss which reduces to a Mahalanobis-type metric learning.

\section{Convolutional neural networks on manifolds}

\subsection{Geodesic convolution}
\begin{figure}[t!]
\vspace{-0.3cm}
\begin{center}
\begin{overpic}[trim=0cm 0cm 0cm 0cm,clip,width=1\linewidth]{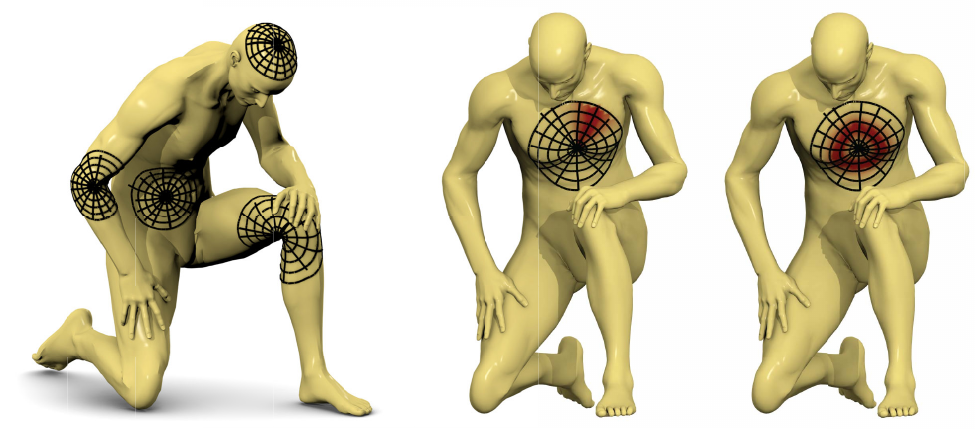}
\put(55,-1.5){\footnotesize $v_\theta$ }
\put(84,-1.5){\footnotesize $v_\rho$ }
\end{overpic}\\
\caption{Construction of local geodesic polar coordinates on a manifold. Left:
examples of local geodesic patches, center and right: example of angular and
radial weights $v_{\theta}$, $v_{\rho}$, respectively (red denotes larger
weights).}
\label{fig:net_on_shape}
\vspace{-0.5cm}
\end{center}
\end{figure}

We introduce a notion of convolution on non-Euclidean domains that follows the `correlation with template' idea by employing a local system of geodesic polar coordinates constructed at point $x$, shown in Figure~\ref{fig:net_on_shape}, to extract patches on the manifold. 
The radial coordinate is constructed as $\rho$-level sets $\{x' : d_X(x,x')=\rho\}$ of the geodesic (shortest path) distance function for $\rho \in [0, \rho_0]$; we call $\rho_0$ the radius of the geodesic disc.
\footnote{
If the radius $\rho_0$ of the geodesic ball $B_{\rho_0}(x) = \{ x' : d_X(x,x') \leq \rho_0\}$ is sufficiently small w.r.t the local convexity radius of the manifold, then the resulting ball is guaranteed to be a topological disc.
} 
Empirically, we see that choosing a sufficiently small $\rho_0 \approx 1\%$ of the geodesic diameter of the shape produces valid topological discs. 
The angular coordinate is constructed as a set of geodesics $\Gamma(x,\theta)$ emanating from $x$ in direction $\theta$; such rays are perpendicular to the geodesic distance level sets.
Note that the choice of the origin of the angular coordinate is arbitrary. 
For boundary points, the procedure is very similar, with the only difference that instead of mapping into a disc we map into a half-disc. 

Let $\Omega(x) \colon B_{\rho_0}(x) \to [0,\rho_0]\times [0,2\pi)$ denote the bijective map from the manifold into the local geodesic polar coordinates $(\rho, \theta)$ around $x$, and let $(D(x)f)(\rho,\theta) = (f\circ \Omega^{-1}(x))(\rho,\theta)$ be the \emph{patch operator} interpolating $f$ in the local coordinates.
We can regard $D(x)f$ as a `patch' on the manifold and use it to define what we term the \emph{geodesic convolution} (GC), 
\begin{equation}
(f\star a)(x) = \sum_{\theta,r} a(\theta+\Delta\theta, r) (D(x) f)(r,\theta), 
\label{eq:geoconv}
\end{equation}  
where $a(\theta, r)$ is a filter applied on the patch. 
Due to angular coordinate ambiguity, the filter can be rotated by arbitrary angle $\Delta\theta$. 

\paragraph*{Patch operator}
Kokkinos et al. \cite{ISC} construct the patch operator as 
\vspace{-0.2cm}
\begin{eqnarray}
(D(x)f)(\rho,\theta) &=& \int_X v_{\rho,\theta}(x,x') f(x') dx',  \\
v_{\rho,\theta}(x,x') &=& \frac{v_\rho(x,x')v_\theta(x,x')}{\int_X  v_\rho(x,x') v_\theta(x,x') dx'}.
\end{eqnarray}
The radial interpolation weight is a Gaussian $v_{\rho}(x,x') \propto e^{-(d_X(x, x')-\rho)^2 / \sigma_\rho^2}$ of the geodesic distance from $x$, centered around $\rho$ (see  Figure~\ref{fig:net_on_shape}, right).
The angular weight is a Gaussian $v_{\theta}(x,x' ) \propto e^{- d^2_{X}(\Gamma(x,\theta), x') / \sigma_\theta^2 }$ of the point-to-set distance $d_{X}(\Gamma(x,\theta), x') = \min_{x'' \in \Gamma(x,\theta)} d_X(x'',x')$ to the geodesic $\Gamma(x,\theta)$ (see  Figure~\ref{fig:net_on_shape}, center). 

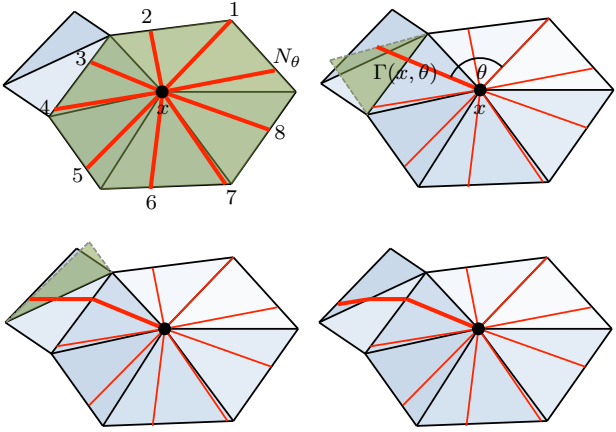
\begin{figure}[t!]
\begin{center}
\begin{overpic}
	[trim=0cm 0cm 0cm 0cm,clip,width=1\linewidth]{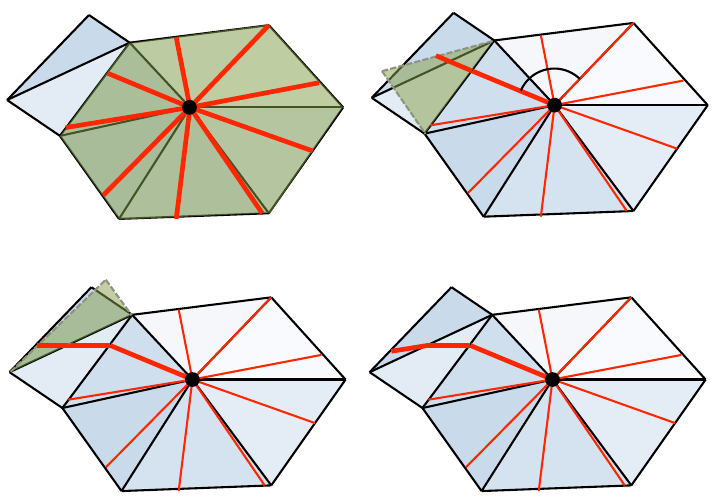}
	\put(25.5,51.5){\footnotesize $x$ }
	\put(37,67){\footnotesize $1$ }
	\put(23,66){\footnotesize $2$ }
	\put(12.5,59.2){\footnotesize $3$ }
	\put(6.75,51.5){\footnotesize $4$ }
	\put(12,40.5){\footnotesize $5$ }
	\put(23.75,36.25){\footnotesize $6$ }
	\put(36.5,37){\footnotesize $7$ }
	\put(44.25,47.5){\footnotesize $8$ }
	\put(44,59.5){\footnotesize $N_\theta$ }
	\put(76,51.5){\footnotesize $x$ }
	\put(76.5,57){\footnotesize $\theta$ }
	\put(60,57){\footnotesize $\Gamma(x,\theta)$ }
\end{overpic}\\
\caption{Construction of local geodesic polar coordinates on a triangular mesh. Shown clock-wise: division of 1-ring of vertex $x_i$ into $N_\theta$ equi-angular bins; propagation of a ray (bold line) by unfolding the respective triangles (marked in green).}
\label{fig:net_on_shape1}
\vspace{-6.5mm}
\end{center}
\end{figure}

\paragraph*{Discrete patch operator}
On triangular meshes, a discrete local system of geodesic polar coordinates has $N_\theta$ angular and $N_\rho$ radial bins.  
Starting with a vertex $i$, we first partition the 1-ring of $i$ by $N_\theta$ rays into equi-angular bins, aligning the first ray with one of the edges (Figure~\ref{fig:net_on_shape1}). 
Next, we propagate the rays into adjacent triangles using an unfolding procedure resembling one used in \cite{kimmel1998computing}, producing poly-lines that form the angular bins (see Figure~\ref{fig:net_on_shape1}).  
Radial bins are created as level sets of the geodesic distance function computed using fast marching \cite{kimmel1998computing}.  

We represent the discrete patch operator as an $N_\theta N_\rho N \times N$ matrix applied to a function defined on the mesh vertices and producing the patches at each vertex. 
The matrix is very sparse since the values of the function at a few nearby vertices only contribute to each local geodesic polar bin. 

\subsection{Geodesic Convolutional Neural Networks}
\begin{figure*}[ht!]
\vspace{-4mm}
\begin{overpic}
	[trim=0cm 0cm 0cm 0cm,clip,width=1\linewidth]{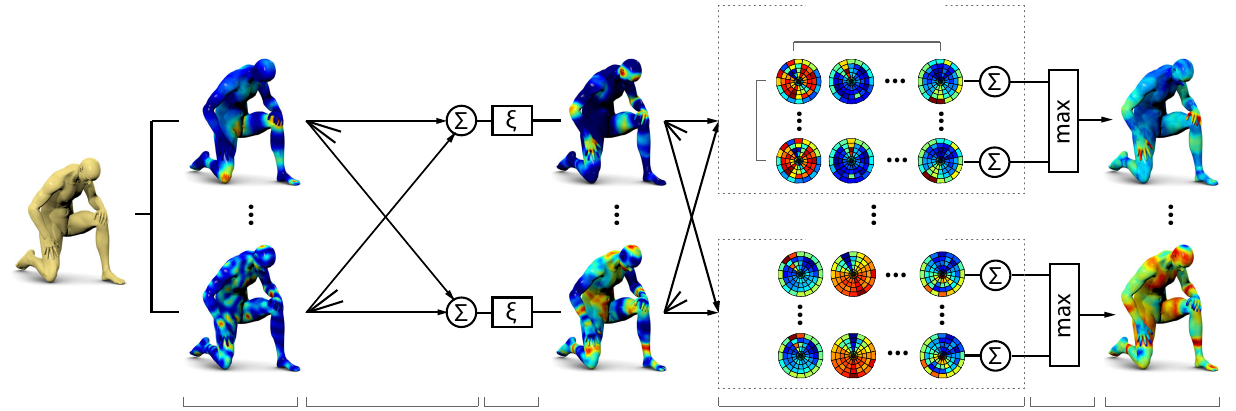}
	\put(15,-1.2){\footnotesize Input $M$-dim}
	\put(31.5,-1.2){\footnotesize LIN}
	\put(39.5,-1.2){\footnotesize ReLU}
	\put(70,-1.2){\footnotesize GC}
	\put(84.4,-1.2){\footnotesize AMP}
	\put(89.5,-1.2){\footnotesize Output $Q$-dim}
	\put(16.5,1.7){\footnotesize $f^\text{in}_M = g_{M}$}	
	\put(17.4,17.05){\footnotesize $f^\text{in}_2 = g_{1}$}
	\put(93.25,1.95){\footnotesize $f^\text{out}_{Q}$}	
	\put(93.5,17.5){\footnotesize $f^\text{out}_{1}$}
	\put(67.2,30.25){\footnotesize $P$ filters}
	\put(59.5,19.55){\rotatebox{90}{\footnotesize $N_\theta$ rotations}}
	\put(66.75,32.25){\footnotesize filter bank 1}
	\put(66.5,13.65){\footnotesize filter bank $Q$}
\end{overpic}
\vspace{-0.95mm}
\caption{
The simple GCNN1 architecture containing one convolutional layer applied to $M=150$-dimensional geometry vectors (input layer) of a human shape, to produce a $Q=16$-dimensional feature descriptor (output layer). 
}	
\vspace{-3mm}
\label{fig:teaser}
\end{figure*}
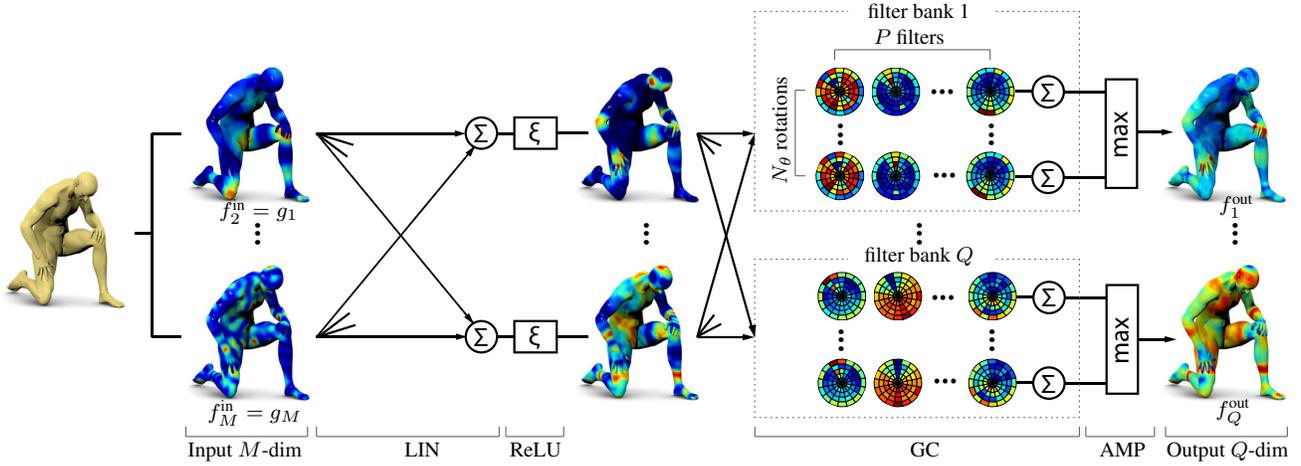

Using the notion of geodesic convolution, we are now ready to extend CNNs to manifolds. 
GCNN consists of several layers that are applied subsequently, i.e. the output of the previous layer is used as the input into the subsequent one (see Figure~\ref{fig:teaser}). 
We distinguish between the following types of layers: 

\paragraph*{Linear (LIN)} layer
typically follows the input layer and precedes the output layer to adjust the 
input and output dimensions by 
means of a linear combination, 
\begin{equation}
f^\text{out}_{q}(x) =
\xi\left( \sum_{p=1}^P w_{qp} f^\text{in}_p(x) \right); 
\quad q = 1,\hdots, Q, 
\label{eq:cnn_fc}
\end{equation}
optionally followed by a non-linear function such as the ReLU, $\xi(t) = \max\{0,t\}$. 

\paragraph*{Geodesic convolution (GC)} layer 
replaces the convolutional layer used in classical Euclidean CNNs. 
Due to the angular coordinate ambiguity, we compute the geodesic convolution result for \emph{all} $N_\theta$ \emph{rotations} of the filters, 
\begin{equation}
\label{eq:gcnn}
f^\mathrm{out}_{\Delta\theta, q}(x) = 
\sum_{p=1}^P (f_p \star a_{\Delta\theta, qp})(x),
\quad q = 1,\hdots, Q, 
\end{equation}
where $a_{\Delta\theta,qp}(\theta,r) = a_{qp}(\theta+\Delta\theta,r)$ are the coefficients of the $p$th filter in the $q$th filter bank rotated by $\Delta\theta = 0, \frac{2\pi}{N_\theta}, \hdots, \frac{2\pi(N_\theta-1)}{N_\theta}$, and the convolution is understood in the sense of~\eqref{eq:geoconv}. 

\paragraph*{Angular max-pooling (AMP)} 
is a fixed layer used in conjunction with the GC layer, that computes the maximum over the filter rotations, 
\begin{equation}
\label{eq:amp}
f^\text{out}_{p}(x) = \max_{\Delta\theta} f^\text{in}_{\Delta\theta, p}(x), \quad p = 1,\hdots, P=Q, 
\end{equation}
where $f^\text{in}_{\Delta\theta, p}$ is the output of the GC layer~\eqref{eq:gcnn}. 

\paragraph*{Fourier transform magnitude (FTM)} layer 
is another fixed layer that applies the patch operator to each input dimension,  followed by Fourier transform w.r.t. the angular coordinate and absolute value, 
\begin{equation}
\label{eq:ftm}
f^\text{out}_{p}(\rho,\omega) = \left | 
\sum_\theta e^{-i\omega \theta }
(D(x) f^\text{in}_{p}(x))(\rho, \theta) 
\right |,
\end{equation}
$p = 1,\hdots, P=Q$. The Fourier transform translates rotational ambiguity into complex phase ambiguity, which is removed by taking the absolute value \cite{kokkinos2008scale,ISC}. 

\paragraph*{Covariance (COV)} layer 
is used in applications such as retrieval where one needs to aggregate the point-wise descriptors and produce a global shape descriptor \cite{tuzel2006region}, 
\begin{equation}
\label{eq:covlayer}
\mathbf{f}^\text{out} = \int_X (\mathbf{f}^\text{in}(x) - \boldsymbol{\mu})(\mathbf{f}^\text{in}(x) - \boldsymbol{\mu})^\top dx,
\end{equation}
where $\mathbf{f}^\text{in}(x) = (f^\text{in}_1(x), \hdots, f^\text{in}_P(x))^\top$ is a $P$-dimensional input vector, $\boldsymbol{\mu} = \int_X \mathbf{f}^\text{in}(x) dx$, and $\mathbf{f}^\text{out}$ is a $P\times P$ matrix column-stacked into a $P^2$-dimensional vector.  

\section{Comparison to previous approaches}
Our approach is perhaps the most natural way of generalizing CNNs to manifolds, where convolutions are performed by sliding a window over the manifold, and local geodesic coordinates are used in place of image `patches'. 
Such patches allow capturing local anisotropic structures. 
Our method is generalizable and unlike spectral approaches does not rely on the approximate invariance of Laplacian eigenfunctions across the shapes. 

\paragraph*{Spectral descriptors} can be obtained as particular configurations of GCNN
applied on geometry vectors input. 
HKS \cite{HKS1} and WKS \cite{WKS} descriptors are obtained by using a fixed LIN layer configured to produce low- or band-pass filters, respectively. 
OSD \cite{LearnDesc} is obtained by using a learnable LIN layer.
Intrinsic shape context \cite{ISC} is obtained by using a fixed LIN layer configured to produce HKS or WKS descriptors, followed by a fixed FTM layer. 

\paragraph*{Spectral nets} \cite{Bruna} are a spectral formulation of CNNs using the notion of generalized (non shift-invariant) convolution that  relies on the analogy between the classical Fourier transform and the Laplace-Beltrami eigenbasis, and the fact that the convolution operator is diagonalized by the Fourier transform. 
The main drawback of this approach is that while it allows to extend CNNs to a
non-Euclidean domain (in particular, the authors considered a graph), it does not generalize across \emph{different} domains; the convolution coefficients are expressed in a domain-dependent basis. 
Another drawback of spectral nets is that they do not use locality. 

\paragraph*{Localized spectral nets} \cite{WFT2015} are an extension of \cite{Bruna} using the Windowed Fourier transform (WFT) \cite{Shumann} on manifolds. 
Due to localization, this method has better generalization abilities, however, it might have problems in the case of strongly non-isometric deformations due to the variability of the Laplacian eigenfunctions. 
Furthermore, while WFT allows capturing local structures, it is \emph{isotropic}, i.e., insensitive to orientations.

\section{Applications}
GCNN model can be thought of as a non-linear hierarchical parametric function $\boldsymbol{\psi}_{\boldsymbol{\Theta}}(\mathbf{F})$, where $\mathbf{F} = (\mathbf{f}(x_1), \hdots, \mathbf{f}(x_N))$ is a $P\times N$ matrix of input features (such as HKS, WKS, geometry vectors, or anything else) at all the points of the mesh, and $\boldsymbol{\Theta}$ denotes the parameters of all the layers. 
Depending on the application in hand, these parameters are learned by minimizing some loss function. We describe three examples of such task-specific losses. 

\paragraph*{Invariant descriptors} 
Applying the GCNN point-wise on some input feature vector $\mathbf{f}(x)$, the output $\boldsymbol{\psi}_{\boldsymbol{\Theta}}(\mathbf{f}(x))$ can be regarded as a dense local descriptor at point $x$. Our goal is to make the output of the network as similar as possible at corresponding points (\emph{positives}) across a collection of shapes, and as dissimilar as possible at non-corresponding points (\emph{negatives}). 
For this purpose, we use a \emph{siamese network} configuration \cite{bromley94,hadsell2006,Simo-SerraTFKM14}, composed of two identical copies of the same GCNN model sharing the same parameterization and fed by pairs of knowingly similar or dissimilar samples, and minimize the \emph{siamese loss}
\begin{eqnarray}
\label{eq:siam}
\ell(\boldsymbol{\Theta}) &=& (1 - \gamma) \sum_{i=1}^{|\mathcal{T}_+|}  \| \boldsymbol{\psi}_{\boldsymbol{\Theta}}(\mathbf{f}_i) - \boldsymbol{\psi}_{\boldsymbol{\Theta}}(\mathbf{f}^+_i) \|^2 \\
&+& \gamma\sum_{i=1}^{|\mathcal{T}_-|}  ( \mu - \| \boldsymbol{\psi}_{\boldsymbol{\Theta}}(\mathbf{f}_i) - \boldsymbol{\psi}_{\boldsymbol{\Theta}}(\mathbf{f}^-_i) \| )_+^2, \nonumber
\end{eqnarray}
where $\gamma \in [0,1]$ is a parameter trading off between the positive and negative losses, $\mu$ is a margin, $(t)_+ = \max \{ 0, t\}$, and $\mathcal{T}_\pm = \{ (\mathbf{f}_i, \mathbf{f}_i^\pm) \}$ denotes the sets of positive and negative pairs, respectively. 

\paragraph*{Shape correspondence}
Finding the correspondence in a collection of shapes can be posed as a labelling problem, where one tries to label each vertex of a given \emph{query} shape $X$ with the index of a corresponding point on some \emph{reference} shape $Y$  \cite{rodoladense}. 
Let $y_1, \hdots, y_{N'}$ be the vertices of $Y$, and let $y_{j_i}$ denote the vertex corresponding to $x_i$ for $i = 1,\hdots, N$. 
GCNN applied point-wise on $X$ is used to produce an $N'$-dimensional vector encoding the probability distribution on $Y$, which acts as a `soft correspondence'. 
The \emph{multinomial regression loss}
\begin{equation}
\label{eq:corresp}
\ell(\boldsymbol{\Theta}) = - \sum_{i=1}^{|\mathcal{T}|} \mathbf{e}_{j_i} \log \boldsymbol{\psi}_{\boldsymbol{\Theta}}(\mathbf{f}_i)
\end{equation}
is minimized on a training set of known correspondence $\mathcal{T} = \{ \mathbf{f}(x_i), j_i \}$ to achieve the optimal correspondence (here $e_i$ is a unit vector with a one at index $i$). 

\paragraph*{Shape retrieval}
In the shape retrieval application, we are interested in producing a global shape descriptor that discriminates between shape classes (note that in a sense this is the converse of invariant descriptors for correspondence, which we wanted to be oblivious to different classes). 
In order to aggregate the local features we use the COV layer in GCNN and regard $\boldsymbol{\psi}_{\boldsymbol{\Theta}}(\mathbf{F})$ as a global shape descriptor. 
Training is done by minimizing the siamese loss, where positives and negatives are shapes from same and different classes, respectively. 

\section{Results}

\begin{figure*}[t!]
\centering
\vspace{-4mm}
\begin{overpic}
	[trim=0cm 0cm 0cm 0cm,clip,width=0.95\linewidth]{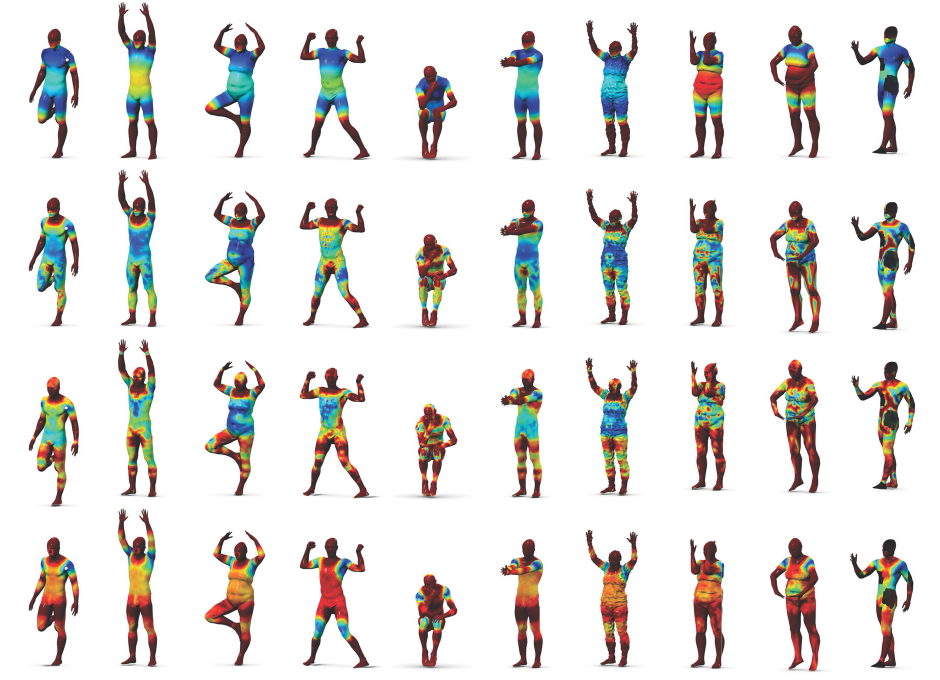}
	\put(42,53.5){\footnotesize Heat kernel signature (HKS)}
	\put(42,35.5){\footnotesize Wave kernel signature (WKS)}
	\put(40.5,17.5){\footnotesize Optimal spectral descriptor (OSD)}
	\put(49,-0.5){\footnotesize GCNN}
\end{overpic}
\vspace{-1mm}
\caption{
Normalized Euclidean distance between the descriptor at a reference point on the shoulder (white sphere) and the descriptors computed at the rest of the points for different transformations (shown left-to-right: near isometric deformations, non-isometric deformations, topological noise, geometric noise, uniform/non-uniform subsampling, missing parts). 
Cold and hot colors represent small and large distances, respectively; distances are saturated at the median value. 
Ideal descriptors would produce a distance map with a sharp minimum at the corresponding point and no spurious local minima at other locations. 
\vspace{-2mm}
}
\label{fig:descs1}
\end{figure*}
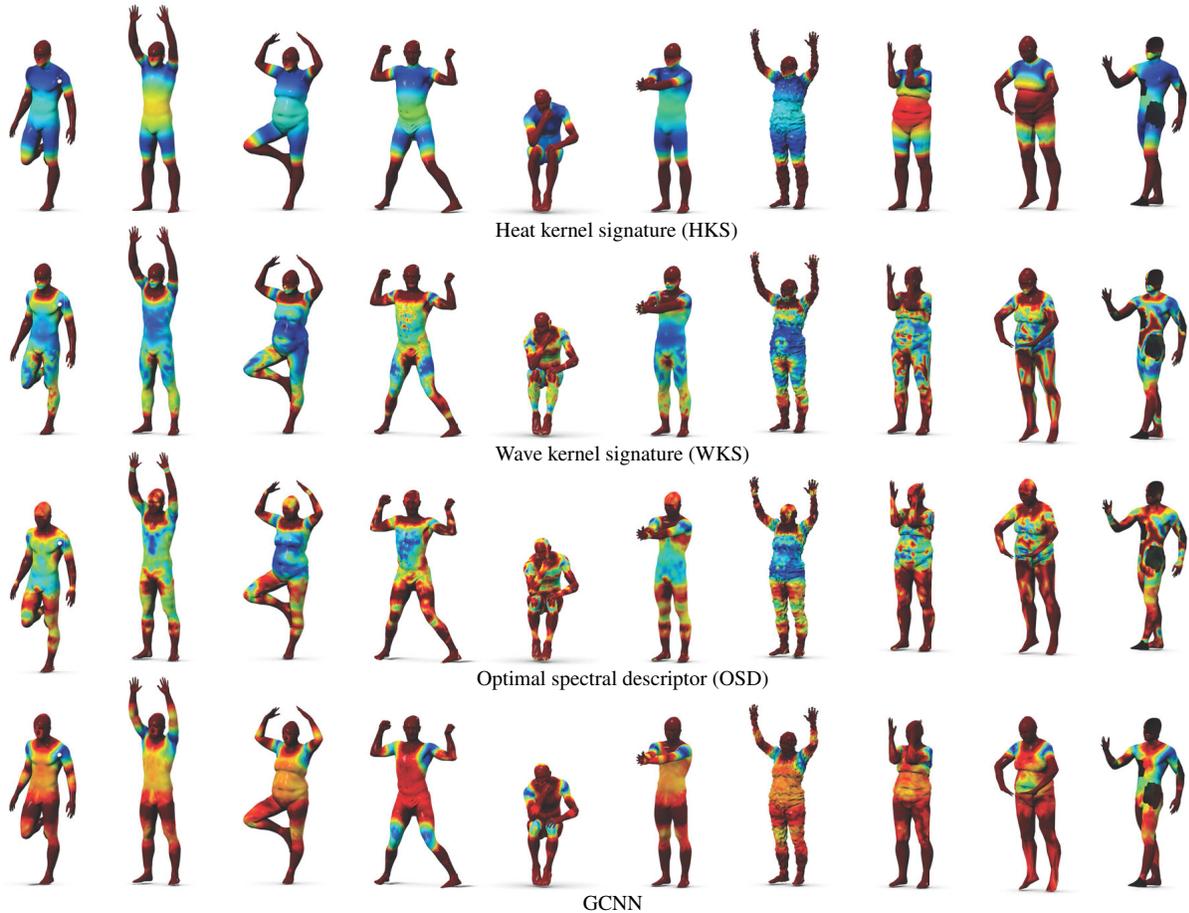

We used the FAUST \cite{FAUST} dataset containing scanned human shapes in different poses and the TOSCA \cite{TOSCA} dataset containing synthetic models of humans in a variety of near-isometric deformations.
The meshes in TOSCA were resampled to $10$K vertices; FAUST shapes contained $6.8$K points.  
All shapes were scaled to unit geodesic diameter. 
GCNN was implemented in Theano \cite{bergstra2010theano}. 
Geodesic patches were generated using the code and settings of \cite{ISC} with $\rho_0 = 1\%$ geodesic diameter. 
Training was performed using the Adadelta stochastic optimization algorithm \cite{zeiler2012} for a maximum of $2.5$K updates. 
Typical training times on FAUST shapes were approximately $30$ and $50$ minutes for one- and two-layer models (GCNN1 and GCNN2, respectively). 
The application of a trained GCNN model to compute feature descriptors was very efficient: $75$K and $45$K vertices/sec for the GCNN1 and GCNN2 models, respectively.  
Training and testing was done on disjoint sets. 
As the input to GCNN, we used $M=150$-dimensional geometry vectors computed according to~\eqref{eq:spline}--\eqref{eq:gendesc_} using B-spline bases.
Laplace-Beltrami operators were discretized using the cotangent formula~\eqref{eq:cotan}; $K=300$ eigenfunctions were computed using MATLAB \texttt{eigs} function. 

\subsection{Intrinsic shape descriptors}
\label{sec:intdesc}
We first used GCNN to produce dense intrinsic pose- and subject-invariant descriptors for human shapes, following nearly-verbatim the experimental setup of \cite{LearnDesc}.
For reference, we compared GCNN to HKS \cite{HKS1}, WKS \cite{WKS}, and OSD \cite{LearnDesc} using the code and settings provided by the respective authors. 
All the descriptors were $Q=16$-dimensional as in \cite{LearnDesc}. 
We used two configurations: GCNN1 ($150$-dim input, LIN16+ReLU, GC16+AMP shown in Figure~\ref{fig:teaser}), and GCNN2 (same as GCNN1 with additional ReLU, FTM, LIN16 layers);  
Training of GCNN was done using the loss~\eqref{eq:siam} with positive and negative sets of vertex pairs generated on the fly.  
On the FAUST dataset, we used subjects $1$--$7$ for training, subject $8$ for validation, and subject $9$--$10$ for testing. 
On TOSCA, we test on all the deformations of the Victoria shape. 

\begin{figure*}[bth]
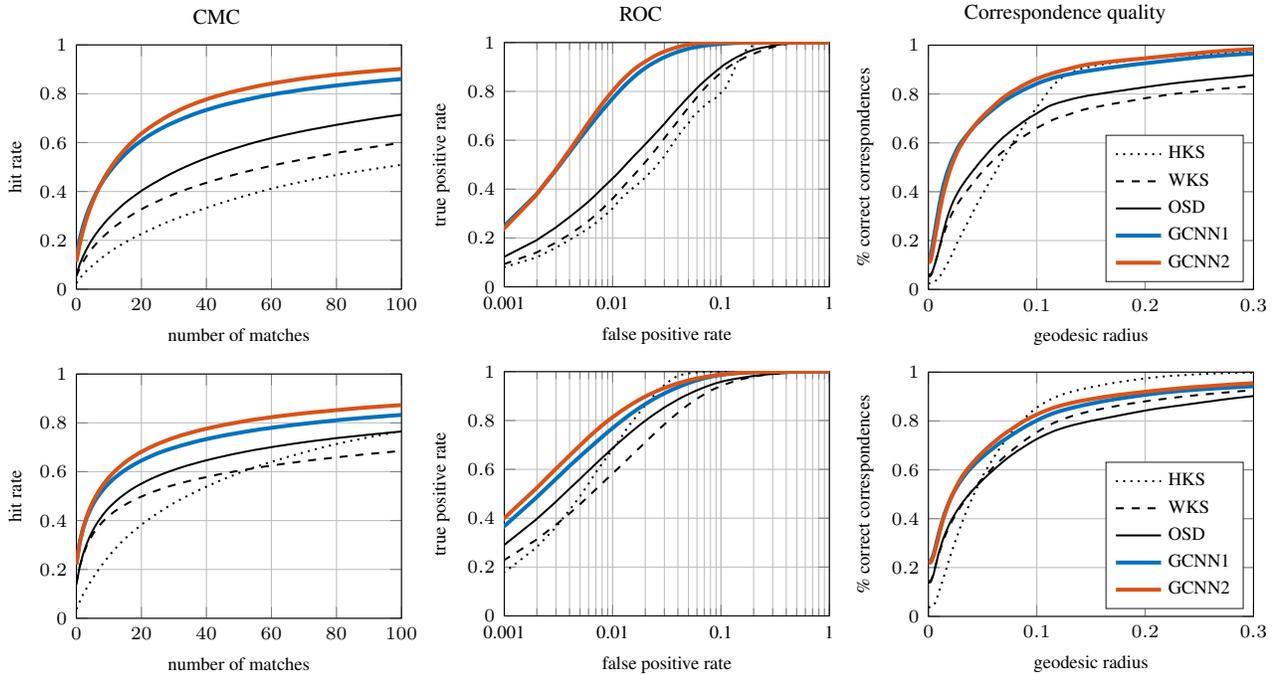

\centering
\begin{minipage}{0.33\textwidth}
\begin{center}
\footnotesize{CMC}
\end{center}
\vspace{-0.3cm}
\setlength\figureheight{3.25cm} 
\setlength\figurewidth{\columnwidth}
\input{./perf_plots/FAUST2FAUST/sym/CMC.tikz}
\end{minipage}
\hspace{-0.3cm}
\begin{minipage}{0.33\textwidth}
\begin{center}
\footnotesize{ROC}
\end{center}
\vspace{-0.3cm}
\setlength\figureheight{3.25cm} 
\setlength\figurewidth{\columnwidth}
\input{./perf_plots/FAUST2FAUST/sym/ROC.tikz}
\end{minipage}
\hspace{-0.3cm}
\begin{minipage}{0.33\textwidth}
\begin{center}
\footnotesize{Correspondence quality}
\end{center}
\vspace{-0.3cm}
\setlength\figureheight{3.25cm} 
\setlength\figurewidth{\columnwidth}
\input{./perf_plots/FAUST2FAUST/sym/Kim.tikz}
\end{minipage}\\
%
\begin{minipage}{0.33\textwidth}
\setlength\figureheight{3.25cm} 
\setlength\figurewidth{\columnwidth}
\input{./perf_plots/FAUST2TOSCA/sym/CMC.tikz}
\end{minipage}
\hspace{-0.3cm}
\begin{minipage}{0.33\textwidth}
\setlength\figureheight{3.25cm} 
\setlength\figurewidth{\columnwidth}
\input{./perf_plots/FAUST2TOSCA/sym/ROC.tikz}
\end{minipage}
\hspace{-0.3cm}
\begin{minipage}{0.33\textwidth}
\setlength\figureheight{3.25cm} 
\setlength\figurewidth{\columnwidth}
\input{./perf_plots/FAUST2TOSCA/sym/Kim.tikz}
\end{minipage}
\caption{
Performance of different descriptors measured using the CMC (left), ROC (center) and Princeton protocol for nearest-neighbor correspondence (right); higher curves correspond to better performance. 
First row show results for GCNN trained and tested on disjoint sets of the FAUST dataset.
Second row shows results for a transfer learning experiment where the net has been trained on FAUST and applied to the TOSCA test set.
GCNN (red and blue curves) significantly outperforms other standard descriptors.
}
\vspace{-3mm}
\label{fig:perf}
\end{figure*}

Figure~\ref{fig:descs1} depicts the Euclidean distance in the descriptor space between the descriptor at a selected point and the rest of the points on the same shape as well as its transformations.
GCNN descriptors manifest both good localization (better than HKS) and are more discriminative (less spurious minima than WKS and OSD), as well as robustness to different kinds of noise, including isometric and non-isometric deformations, geometric and topological noise, different sampling, and missing parts. 

Quantitative descriptor evaluation was done using three criteria: \emph{cumulative match characteristic} (CMC), \emph{receiver operator characteristic} (ROC), and the \emph{Princeton protocol} \cite{Blended}. 
The CMC evaluates the probability of a correct correspondence among the $k$ nearest neighbors in the descriptor space. 
The ROC measures the percentage of positives and negatives pairs falling below various thresholds of their distance in the descriptor space (\emph{true positive} and \emph{negative rates}, respectively).
The Princeton protocol counting the percentage of nearest-neighbor matches that are at most $r$-geodesically distant from the groundtruth correspondence. 
Figure~\ref{fig:perf} (first row) shows the performance of different descriptors on the FAUST dataset. 
We observe that GCNN descriptors significantly outperform other descriptors, and that the more complex model (GCNN2) further boosts performance. 
In order to test the generalization capability of the learned descriptors, we
applied OSD and GCNN learned on the FAUST dataset to TOSCA shapes (Figure \ref{fig:perf}, second row). 
We see that the learned model transfers well to a new dataset. 

\subsection{Shape correspondence}
\begin{figure}[t!]
\begin{center}
\setlength\figureheight{3.25cm} 
\setlength\figurewidth{0.8\columnwidth}
\input{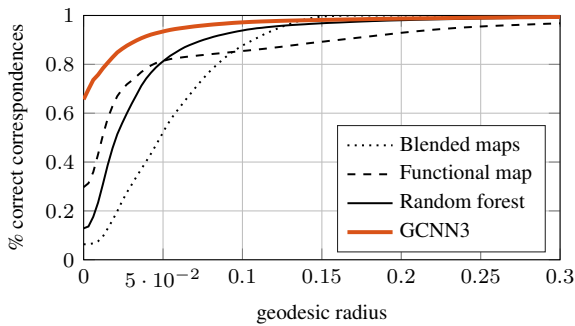}
\end{center}\vspace{-5mm}
\caption{
Performance of shape correspondence on the FAUST dataset evaluated using the Princeton benchmark. 
Higher curve corresponds to better performance.
}\vspace{-3mm}
\label{fig:corr1}
\end{figure}

\begin{figure}[t!]
\begin{center}
\begin{overpic}
	[trim=0cm 0cm 0cm 0cm,clip,width=0.9\linewidth]{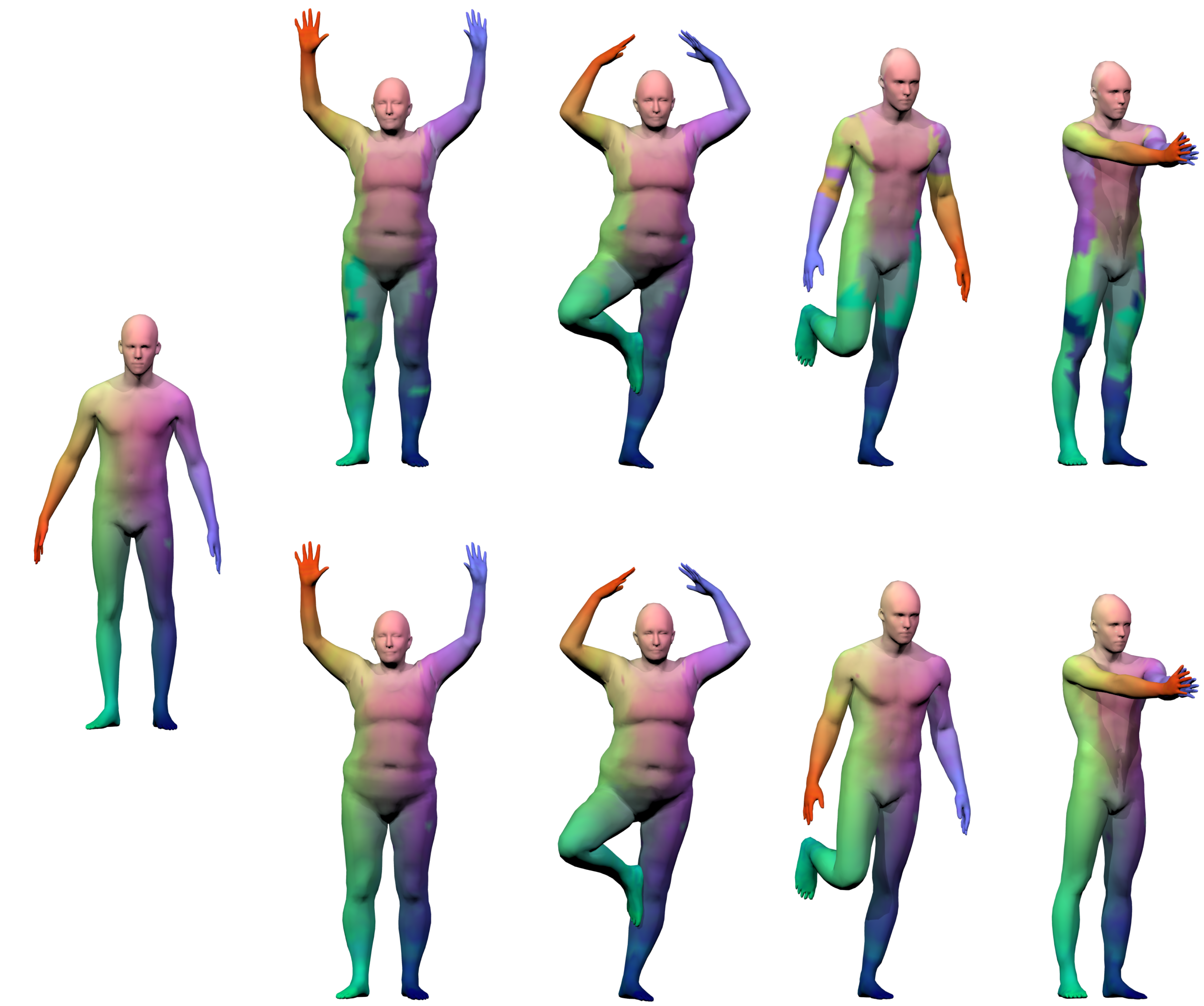}
	\put(54,40){\footnotesize Random Forest }
	\put(58,-2.5){\footnotesize GCNN }
\end{overpic}\\
\end{center}\vspace{-2.5mm}
\caption{
Example of correspondence obtained with GCNN (bottom) and random forest (top). Similar colors encode corresponding points.
}
\vspace{-3mm}
\label{fig:corr2}
\end{figure}

To show the application of GCNN for computing intrinsic correspondence, we reproduced the experiment of Rodol{\`a} et al. \cite{rodoladense} on the FAUST dataset, replacing their random forest with a GCNN architecture GCNN3 containing three convolutional layers (input: $150$-dimensional geometry vectors, LIN16+ReLU, GC32+AMP+ReLU, GC64+AMP+ReLU, GC128+AMP+ReLU, LIN256, LIN6890). 
Zeroth FAUST shape containing $N'=6890$ vertices was used as reference; for each point on the query shape, the output of GCNN representing the soft correspondence as an $6890$-dimensional vector was converted into a point correspondence by taking the maximum. 
Training was done by minimizing the loss~(\ref{eq:corresp}); training and test sets were as in the previous experiment.  
Figure~\ref{fig:corr1} shows the performance of our method evaluated using the Princeton benchmark, and Figure~\ref{fig:corr2} shows correspondence examples where colors are transferred using raw point-wise correspondence in input to the functional maps algorithm.
GCNN shows significantly better performance than previous methods \cite{Blended,FuncMaps,rodoladense}. 


\subsection{Shape retrieval}
\begin{figure}[t!]
\begin{center}
\setlength\figureheight{3.25cm} 
\setlength\figurewidth{0.8\columnwidth}
%
%
\definecolor{mycolor1}{rgb}{0.00000,0.44700,0.74100}%
\definecolor{mycolor2}{rgb}{0.85000,0.32500,0.09800}%
\begin{tikzpicture}

\tikzstyle{every node}=[font=\footnotesize]

\begin{axis}[%
compat=newest,
width=0.95092\figurewidth,
height=\figureheight,
at={(0\figurewidth,0\figureheight)},
scale only axis,
xmin=0,
xmax=1,
xlabel={Recall},
ymin=0,
ymax=1,
ylabel={Precision},
legend style={at={(0.03,0.03)},anchor=south west,legend cell align=left,align=left,draw=white!15!black}
]
\addplot [color=black,dotted,line width=0.75pt]
  table[row sep=crcr]{%
0	1\\
0.17	0.680000000032\\
0.245	0.490000000013\\
0.315	0.420000000004\\
0.365	0.365000000002\\
0.405	0.3240000000016\\
0.445	0.296666666667333\\
0.485	0.277142857143429\\
0.505	0.2525000000005\\
0.515	0.228888888889333\\
0.545	0.2180000000004\\
0.56	0.203636363636727\\
0.575	0.191666666667\\
0.59	0.181538461538769\\
0.62	0.177142857143143\\
0.635	0.1693333333336\\
0.66	0.165000000000125\\
0.68	0.160000000000118\\
0.685	0.152222222222222\\
0.7	0.147368421052632\\
0.705	0.141\\
0.72	0.137142857142857\\
0.735	0.133636363636364\\
0.745	0.129565217391304\\
0.76	0.126666666666667\\
0.77	0.1232\\
0.775	0.119230769230769\\
0.785	0.116296296296296\\
0.79	0.112857142857143\\
0.795	0.109655172413793\\
0.82	0.109333333333333\\
0.835	0.107741935483871\\
0.84	0.105\\
0.845	0.102424242424242\\
0.86	0.101176470588235\\
0.865	0.0988571428571428\\
0.88	0.0977777777777778\\
0.89	0.0962162162162161\\
0.895	0.0942105263157894\\
0.895	0.0917948717948718\\
0.905	0.0905000000000001\\
0.925	0.0902439024390244\\
0.935	0.0890476190476191\\
0.94	0.0874418604651163\\
0.95	0.0863636363636364\\
0.97	0.0862222222222222\\
0.975	0.0847826086956521\\
0.975	0.0829787234042553\\
0.99	0.0825000000000001\\
1	0.0816326530612245\\
};
\addlegendentry{HKS};

\addplot [color=black,dashed,line width=0.75pt]
  table[row sep=crcr]{%
0	1\\
0.23	0.920000000008\\
0.41	0.820000000002\\
0.5	0.666666666667334\\
0.54	0.5400000000005\\
0.6	0.48\\
0.66	0.44\\
0.695	0.397142857142857\\
0.75	0.375\\
0.78	0.346666666666667\\
0.8	0.32\\
0.815	0.296363636363636\\
0.835	0.278333333333333\\
0.845	0.26\\
0.86	0.245714285714286\\
0.87	0.232\\
0.9	0.225\\
0.91	0.214117647058823\\
0.91	0.202222222222222\\
0.93	0.19578947368421\\
0.935	0.187\\
0.94	0.179047619047619\\
0.94	0.170909090909091\\
0.945	0.164347826086956\\
0.96	0.16\\
0.97	0.1552\\
0.975	0.15\\
0.975	0.144444444444444\\
0.975	0.139285714285714\\
0.975	0.13448275862069\\
0.98	0.130666666666667\\
0.98	0.126451612903226\\
0.98	0.1225\\
0.985	0.119393939393939\\
0.99	0.116470588235294\\
0.995	0.113714285714286\\
1	0.111111111111111\\
1	0.108108108108108\\
1	0.105263157894737\\
1	0.102564102564103\\
1	0.1\\
1	0.0975609756097561\\
1	0.0952380952380952\\
1	0.0930232558139536\\
1	0.0909090909090909\\
1	0.0888888888888889\\
1	0.0869565217391304\\
1	0.0851063829787234\\
1	0.0833333333333334\\
1	0.0816326530612245\\
};
\addlegendentry{WKS};

\addplot [color=black,solid,line width=0.75pt]
  table[row sep=crcr]{%
0	1\\
0.195	0.780000000022\\
0.305	0.610000000009\\
0.36	0.480000000006\\
0.39	0.390000000004\\
0.425	0.3400000000024\\
0.445	0.296666666668333\\
0.47	0.268571428572571\\
0.5	0.2500000000005\\
0.54	0.240000000000444\\
0.555	0.2220000000004\\
0.575	0.209090909091273\\
0.595	0.198333333333667\\
0.625	0.192307692308\\
0.645	0.184285714286\\
0.68	0.181333333333467\\
0.69	0.172500000000125\\
0.7	0.164705882353059\\
0.72	0.160000000000111\\
0.745	0.156842105263263\\
0.76	0.1520000000001\\
0.775	0.147619047619143\\
0.79	0.143636363636455\\
0.8	0.139130434782696\\
0.8	0.133333333333417\\
0.805	0.12880000000008\\
0.82	0.126153846153923\\
0.84	0.124444444444518\\
0.855	0.122142857142857\\
0.855	0.117931034482759\\
0.87	0.116\\
0.88	0.113548387096774\\
0.89	0.11125\\
0.9	0.109090909090909\\
0.91	0.107058823529412\\
0.91	0.104\\
0.92	0.102222222222222\\
0.925	0.0999999999999999\\
0.925	0.0973684210526315\\
0.93	0.0953846153846154\\
0.93	0.093\\
0.94	0.0917073170731708\\
0.955	0.090952380952381\\
0.965	0.0897674418604652\\
0.97	0.0881818181818181\\
0.98	0.0871111111111111\\
0.99	0.0860869565217391\\
0.995	0.0846808510638298\\
0.995	0.0829166666666667\\
1	0.0816326530612245\\
};
\addlegendentry{OSD};

\addplot [color=mycolor2,solid,line width=1.5pt]
  table[row sep=crcr]{%
0	1\\
0.24	0.960000000004\\
0.465	0.930000000001\\
0.67	0.893333333334\\
0.785	0.7850000000005\\
0.825	0.6600000000004\\
0.875	0.583333333333334\\
0.91	0.52\\
0.93	0.465\\
0.96	0.426666666666666\\
0.98	0.392\\
0.99	0.36\\
1	0.333333333333333\\
1	0.307692307692308\\
1	0.285714285714286\\
1	0.266666666666667\\
1	0.25\\
1	0.235294117647059\\
1	0.222222222222222\\
1	0.210526315789474\\
1	0.2\\
1	0.19047619047619\\
1	0.181818181818182\\
1	0.173913043478261\\
1	0.166666666666667\\
1	0.16\\
1	0.153846153846154\\
1	0.148148148148148\\
1	0.142857142857143\\
1	0.137931034482759\\
1	0.133333333333333\\
1	0.129032258064516\\
1	0.125\\
1	0.121212121212121\\
1	0.117647058823529\\
1	0.114285714285714\\
1	0.111111111111111\\
1	0.108108108108108\\
1	0.105263157894737\\
1	0.102564102564103\\
1	0.1\\
1	0.0975609756097561\\
1	0.0952380952380952\\
1	0.0930232558139536\\
1	0.0909090909090909\\
1	0.0888888888888889\\
1	0.0869565217391304\\
1	0.0851063829787234\\
1	0.0833333333333334\\
1	0.0816326530612245\\
};
\addlegendentry{GCNN}; 

\end{axis}
\end{tikzpicture}%
\end{center}
\vspace{-5mm}
\caption{
Performance (in terms of Precision-Recall) of shape retrieval on the FAUST dataset using different descriptors. Higher curve corresponds to better performance.
}
\label{fig:prcurve}
\end{figure}
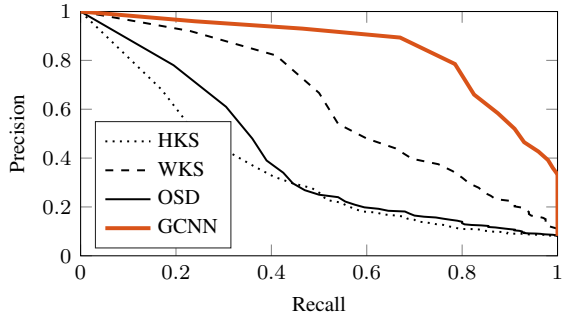

In our final experiment, we performed pose-invariant shape retrieval on the FAUST dataset. 
This is a hard fine-grained classification problem since some of the human subjects look nearly identical. 
We used a GCNN architecture with one convolutional layer (input: $16$-dimensional HKS descriptors, LIN8, GC8+AMP, COV), producing a  
$64$-dimensional output used as the global shape descriptor. 
Training set consisted of five poses per subject (a total of $50$ shapes); testing was performed on the $50$ remaining shapes in a leave-one-out fashion. 
Evaluation was done in terms of \emph{precision} (percentage of retrieved shapes matching the query class) and \emph{recall} (percentage of shapes from the query class that is retrieved).  
Figure~\ref{fig:prcurve} shows the PR curve. 
For comparison, we show the performance of other descriptors (HKS, WKS, and OSD) aggregated into a global covariance shape descriptor. GCNN outperforms significantly all other methods. 

\section{Conclusions}

We presented GCNN, a generalization of CNNs allowing to learn hierarchical task-specific features on non-Euclidean manifolds for applications such as shape correspondence or retrieval.
Our model is very generic and flexible, and can be made arbitrarily complex by stacking multiple layers.
Applying GCNN on other shape representations such as point clouds could be achieved by modifying the local geodesic charting procedure.
Though in this paper we used intrinsic spectral properties of the shape as the the input to the network, GCNN can be applied on any function defined on the manifold, and it would be particularly natural to use it to construct descriptors of textured surfaces.

\subsubsection*{Acknowledgments}
This work was supported in part by the ERC Starting Grant No. 307047 (COMET).


{\small
\bibliographystyle{ieee}
\bibliography{biblio}

\begin{thebibliography}{10}\itemsep=-1pt

\bibitem{WKS}
M.~Aubry, U.~Schlickewei, and D.~Cremers.
\newblock {T}he wave kernel signature: {A} quantum mechanical approach to shape
  analysis.
\newblock In {\em {P}roc. {ICCV}}, 2011.

\bibitem{belongie2000shape}
S.~Belongie, J.~Malik, and J.~Puzicha.
\newblock Shape context: A new descriptor for shape matching and object
  recognition.
\newblock In {\em Proc. NIPS}, 2000.

\bibitem{bergstra2010theano}
J.~Bergstra et~al.
\newblock Theano: a {CPU} and {GPU} math expression compiler.
\newblock In {\em Proc. {SciPy}}, June 2010.

\bibitem{FAUST}
F.~Bogo et~al.
\newblock {FAUST}: {D}ataset and evaluation for {3D} mesh registration.
\newblock In {\em {P}roc. {CVPR}}, 2014.

\bibitem{WFT2015}
D.~Boscaini et~al.
\newblock Learning class-specific descriptors for deformable shapes using
  localized spectral convolutional networks.
\newblock {\em CGF}, 34(5):13--23, 2015.

\bibitem{bromley94}
J.~Bromley et~al.
\newblock Signature verification using a ``{S}iamese'' time delay neural
  network.
\newblock In {\em Proc. NIPS}. 1994.

\bibitem{TOSCA}
A.~M. Bronstein, M.~M. Bronstein, and R.~Kimmel.
\newblock {\em {N}umerical Geometry of Non-Rigid Shapes}.
\newblock Springer, 2008.

\bibitem{bronstein2011shape}
A.~M. Bronstein et~al.
\newblock Shape {G}oogle: Geometric words and expressions for invariant shape
  retrieval.
\newblock {\em TOG}, 30(1):1--20, 2011.

\bibitem{Bruna}
J.~Bruna et~al.
\newblock {S}pectral networks and locally connected networks on graphs.
\newblock In {\em {P}roc. {ICLR}}, 2014.

\bibitem{Ciresan:2012f}
D.~C. Ciresan et~al.
\newblock Deep neural networks segment neuronal membranes in electron
  microscopy images.
\newblock In {\em Proc. NIPS}, 2012.

\bibitem{coifman2006diffusion}
R.~R. Coifman and S.~Lafon.
\newblock Diffusion maps.
\newblock {\em Applied and Comp. Harmonic Analysis}, 21(1):5--30, 2006.

\bibitem{cormansupervised}
{\'E}.~Corman, M.~Ovsjanikov, and A.~Chambolle.
\newblock Supervised descriptor learning for non-rigid shape matching, 2014.

\bibitem{dalal2005histograms}
N.~Dalal and B.~Triggs.
\newblock Histograms of oriented gradients for human detection.
\newblock In {\em Proc. CVPR}, 2005.

\bibitem{digne2010level}
J.~Digne et~al.
\newblock The level set tree on meshes.
\newblock In {\em Proc. 3DPVT}, 2010.

\bibitem{elad2003bending}
A.~Elad and R.~Kimmel.
\newblock On bending invariant signatures for surfaces.
\newblock {\em PAMI}, 25(10):1285--1295, 2003.

\bibitem{FangGISDDGHMPZZ14}
H.~Fang et~al.
\newblock From captions to visual concepts and back.
\newblock {\em arXiv:1411.4952}, 2014.

\bibitem{fukushima1980neocognitron}
K.~Fukushima.
\newblock Neocognitron: A self-organizing neural network model for a mechanism
  of pattern recognition unaffected by shift in position.
\newblock {\em Biological Cybernetics}, 36(4):193--202, 1980.

\bibitem{hadsell2006}
R.~Hadsell, S.~Chopra, and Y.~LeCun.
\newblock Dimensionality reduction by learning an invariant mapping.
\newblock In {\em Proc. CVPR}, 2006.

\bibitem{johnson1999using}
A.~E. Johnson and M.~Hebert.
\newblock Using spin images for efficient object recognition in cluttered {3D}
  scenes.
\newblock {\em PAMI}, 21(5):433--449, 1999.

\bibitem{kanezakilearning}
A.~Kanezaki et~al.
\newblock Learning similarities for rigid and non-rigid object detection.
\newblock In {\em Proc. BMVC}, 2014.

\bibitem{KarpathyF14}
A.~Karpathy and L.~Fei{-}Fei.
\newblock Deep visual-semantic alignments for generating image descriptions.
\newblock {\em arXiv:1412.2306}, 2014.

\bibitem{Blended}
V.~G. Kim, Y.~Lipman, and T.~Funkhouser.
\newblock {B}lended intrinsic maps.
\newblock {\em TOG}, 30(4):1--12, 2011.

\bibitem{kimmel1998computing}
R.~Kimmel and J.~A. Sethian.
\newblock Computing geodesic paths on manifolds.
\newblock {\em PNAS}, 95(15):8431--8435, 1998.

\bibitem{ISC}
I.~Kokkinos et~al.
\newblock {I}ntrinsic shape context descriptors for deformable shapes.
\newblock In {\em {P}roc. {CVPR}}, 2012.

\bibitem{kokkinos2008scale}
I.~Kokkinos and A.~Yuille.
\newblock Scale invariance without scale selection.
\newblock In {\em Proc. CVPR}, 2008.

\bibitem{Krizhevsky:2012}
A.~Krizhevsky, I.~Sutskever, and G.~E. Hinton.
\newblock Image{N}et classification with deep convolutional neural networks.
\newblock In {\em Proc. NIPS}, 2012.

\bibitem{lecun1989backpropagation}
Y.~LeCun et~al.
\newblock Backpropagation applied to handwritten zip code recognition.
\newblock {\em Neural Comp.}, 1(4):541--551, 1989.

\bibitem{levy2006laplace}
B.~L{\'e}vy.
\newblock Laplace-{B}eltrami eigenfunctions towards an algorithm that
  ``understands'' geometry.
\newblock In {\em Proc. SMI}, 2006.

\bibitem{LearnDesc}
R.~Litman and A.~M. Bronstein.
\newblock Learning spectral descriptors for deformable shape correspondence.
\newblock {\em PAMI}, 36(1):170--180, 2014.

\bibitem{litman2014supervised}
R.~Litman et~al.
\newblock Supervised learning of bag-of-features shape descriptors using sparse
  coding.
\newblock {\em CGF}, 33(5):127--136, 2014.

\bibitem{lowe2004distinctive}
D.~G. Lowe.
\newblock Distinctive image features from scale-invariant keypoints.
\newblock {\em IJCV}, 60(2):91--110, 2004.

\bibitem{manay2006integral}
S.~Manay et~al.
\newblock Integral invariants for shape matching.
\newblock {\em PAMI}, 28(10):1602--1618, 2006.

\bibitem{matas2004robust}
J.~Matas et~al.
\newblock Robust wide-baseline stereo from maximally stable extremal regions.
\newblock {\em IVC}, 22(10):761--767, 2004.

\bibitem{osada2002shape}
R.~Osada et~al.
\newblock Shape distributions.
\newblock {\em TOG}, 21(4):807--832, 2002.

\bibitem{FuncMaps}
M.~Ovsjanikov et~al.
\newblock Functional maps: a flexible representation of maps between shapes.
\newblock {\em TOG}, 31(4):1--11, 2012.

\bibitem{Pinkall1993}
U.~Pinkall and K.~Polthier.
\newblock Computing discrete minimal surfaces and their conjugates.
\newblock {\em Experimental Mathematics}, 2(1):15--36, 1993.

\bibitem{rodoladense}
E.~Rodol\`{a} et~al.
\newblock Dense non-rigid shape correspondence using random forests.
\newblock In {\em Proc. CVPR}, 2014.

\bibitem{Sermanet14}
P.~Sermanet et~al.
\newblock Over{F}eat: Integrated recognition, localization and detection using
  convolutional networks.
\newblock In {\em Proc. ICLR}, 2014.

\bibitem{shotton2013real}
J.~Shotton et~al.
\newblock Real-time human pose recognition in parts from single depth images.
\newblock {\em Comm. ACM}, 56(1):116--124, 2013.

\bibitem{Shumann}
D.~I. Shuman, B.~Ricaud, and P.~Vandergheynst.
\newblock Vertex-frequency analysis on graphs.
\newblock {\em arXiv:1307.5708}, 2013.

\bibitem{Simo-SerraTFKM14}
E.~Simo{-}Serra et~al.
\newblock Fracking deep convolutional image descriptors.
\newblock {\em arXiv:1412.6537}, 2014.

\bibitem{Simonyan14c}
K.~Simonyan and A.~Zisserman.
\newblock Very deep convolutional networks for large-scale image recognition.
\newblock {\em arXiv:1409.1556}, 2014.

\bibitem{skraba2010persistence}
P.~Skraba et~al.
\newblock Persistence-based segmentation of deformable shapes.
\newblock In {\em Proc. NORDIA}, 2010.

\bibitem{kalogerakis2015}
H.~Su et~al.
\newblock Multi-view convolutional neural networks for {3D} shape recognition.
\newblock In {\em Proc. ICCV}, 2015.

\bibitem{HKS1}
J.~Sun, M.~Ovsjanikov, and L.~J. Guibas.
\newblock A concise and provably informative multi-scale signature based on
  heat diffusion.
\newblock {\em CGF}, 28(5):1383--1392, 2009.

\bibitem{tuzel2006region}
O.~Tuzel, F.~Porikli, and P.~Meer.
\newblock Region covariance: A fast descriptor for detection and
  classification.
\newblock In {\em Proc. ECCV}, 2006.

\bibitem{windheuseroptimal}
T.~Windheuser et~al.
\newblock Optimal intrinsic descriptors for non-rigid shape analysis.
\newblock In {\em Proc. BMVC}, 2014.

\bibitem{wu2015}
Z.~Wu et~al.
\newblock {3D ShapeNets}: A deep representation for volumetric shape modeling.
\newblock In {\em Proc. CVPR}, 2015.

\bibitem{zaharescu2009surface}
A.~Zaharescu et~al.
\newblock Surface feature detection and description with applications to mesh
  matching.
\newblock In {\em Proc. CVPR}, 2009.

\bibitem{zeiler2012}
M.~D. Zeiler.
\newblock {ADADELTA:} {A}n adaptive learning rate method.
\newblock {\em arXiv:1212.5701}, 2012.

\end{thebibliography}
}

\end{document}